\documentclass{article}
\usepackage{absspconf,amsmath,epsfig}
\usepackage{multicol}
\usepackage{floatrow}
\usepackage{balance}
\usepackage{subcaption}
\usepackage{hyperref}
\usepackage{float}

\title{Quantitative Analysis of Primary Attribution Explainable Artificial Intelligence Methods for Remote Sensing Image Classification}
\name{Akshatha Mohan and Joshua Peeples \thanks{Portions of this research were conducted with the advanced computing resources provided by Texas A\&M High Performance Research Computing.}}
\address{Department of Electrical and Computer Engineering, Texas A\&M University, College Station, TX, USA}

\begin{document}
%
\maketitle

\begin{abstract}
We present a comprehensive analysis of quantitatively evaluating explainable artificial intelligence (XAI) techniques for remote sensing image classification. Our approach leverages state-of-the-art machine learning approaches to perform remote sensing image classification across multiple modalities. We investigate the results of the models qualitatively through XAI methods. Additionally, we compare the XAI methods quantitatively through various categories of desired properties. Through our analysis, we offer insights and recommendations for selecting the most appropriate XAI method(s) to gain a deeper understanding of the models' decision-making processes. The code for this work is publicly available 
\footnote{\url{https://github.com/Peeples-Lab/XAI\_Analysis}}.

\end{abstract}

\section{Introduction} Remote sensing is valuable in several real-world applications such as defense, agriculture and environmental monitoring \cite{richards2022remote}. Artificial intelligence (AI) and machine learning (ML) have assisted in automating analysis. Despite the success of AI/ML, these methods (particularly deep learning approaches) are often viewed as a ``black box" and this can be detrimental in practice by leading to possible mistrust, bias, and ethical concerns \cite{xu2022ai}. To combat these negative aspects, explainable AI (XAI) methods have been developed to elucidate the decisions that are made by these models. XAI approaches provide qualitative insights into the model, but the choice of which method can be difficult in practice \cite{hedstrom2023quantus,bommer2023finding}.

Throughout remote sensing, selecting the most appropriate XAI method(s) to understand the models is important. To investigate the selection of XAI method(s), we perform a benchmark study to evaluate state-of-the-art models such as convolutional neural networks (CNNs) \cite{liu2022convnet}, transformers \cite{khan2022transformers}, and Focal Modulation Networks (FocalNets) \cite{yang2022focal} using XAI methods and metrics. To our knowledge, this work is the first to quantitatively assess XAI methods for different remote sensing modalities across multiple categories of XAI metrics. Previous work only focused on a subset of these metrics (\textit{i.e.}, robustness and complexity) \cite{kakogeorgiou2021evaluating} or only accessed performance without XAI metrics \cite{papoutsis2023benchmarking}; however, a holistic evaluation of each approach is needed. Through our work, we aim to provide insight and develop a systematic approach to assess performance and explainability for remote sensing image classification.

\begin{figure}[t]
    \centering
    \includegraphics[width=1\linewidth]{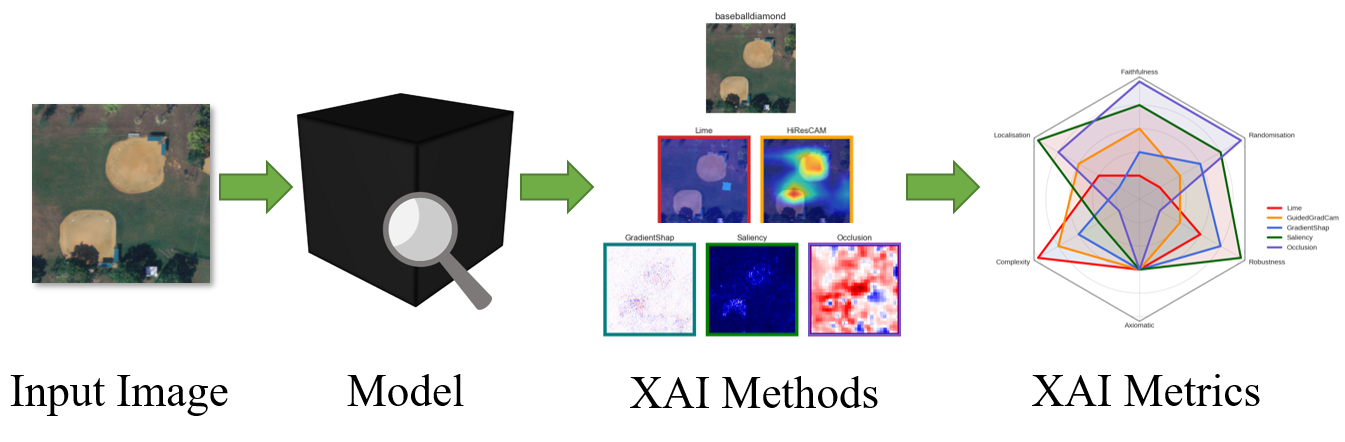}
\caption{Overall experimental pipeline for study.} 
\label{fig:Method}
\end{figure}

\section{Method}
\subsection{XAI Approaches}
\label{sect:XAI_methods}
The overall approach for our work is shown in Figure \ref{fig:Method}. The first step is to train and validate three types of state-of-the-art models: ConvNeXt \cite{liu2022convnet}, vision transformers (ViT) \cite{dosovitskiyimage}, and FocalNets \cite{yang2022focal}. Each model is evaluated using XAI approaches focused on primary attribution (\textit{i.e.}, influence of input data on model output) \cite{kokhlikyan2020captum}. The study focused on five common primary attribution XAI approaches: High-Resolution Class Activation Mapping (HiResCAM) \cite{draelos2021use}, Local Interpretable Model-agnostic Explanations (LIME), Gradient SHapley Additive exPlanations (GradSHAP) \cite{lundberg2017unified}, Saliency maps \cite{simonyan2014deep}, and  Occlusion \cite{zhou2016learning}. To select the best XAI approach, we assess the performance across the six categories of XAI metrics discussed in Section \ref{sect:XAI_metrics}.

\subsection{XAI metrics}
\label{sect:XAI_metrics}
 XAI methods shed light on the predictions of deep neural networks (DNNs). However, XAI methods are often missing ground truth explanations which complicate their evaluation and validation, subsequently compounding the choice of the XAI method \cite{hedstrom2023quantus}. XAI methods can be evaluated across desired explanation properties, namely, robustness, faithfulness, randomization, complexity, localization, and axiomatic \cite{hedstrom2023quantus}. The metrics used in this study were Max Sensitivity (robustness), Faithfulness Correlation (faithfulness), Relevance Rank Accuracy (localization), Sparseness (complexity), Model Parameter Randomization (randomization), and Completeness (axiomatic). 

\subsection{Implementation Details}
\label{sect:Implementation}
\begin{figure}[htb]
    \centering
    \includegraphics[width=1\linewidth]{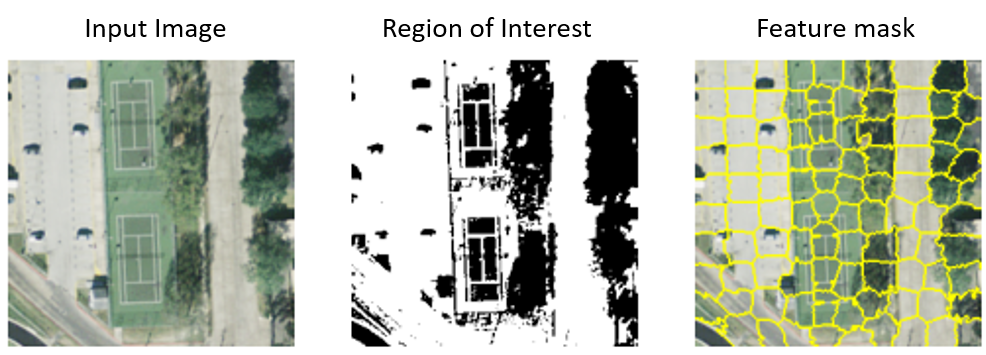}
    \caption{An example binary threshold mask and feature mask using superpixels \cite{achanta2012slic} generated from an image from UCMerced. Localization XAI metrics need a binary mask for the region of interest and a feature mask is needed for LIME to improve computation.}
    \label{fig:implementation details}
\end{figure}

Localization metrics require a region of interest (ROI) that can be in the form of a segmentation mask or bounding box \cite{hedstrom2023quantus,bommer2023finding}. The metric captures the relevance between the ROI and the highly relevant pixels that the model used for the prediction. The datasets used in this work did not have ROIs readily available. To resolve this problem, Otsu's threshold method \cite{otsu1979threshold} was applied on a per-channel basis (only applied to the magnitude for the synthetic aperture radar or SAR). The resulting mask from each channel was then combined through elementwise multiplication to generate a mask for the image.
To improve the efficiency of LIME, a feature mask was generated to reduce computation. Superpixel segmentation \cite{achanta2012slic} was used to produce a feature mask. An example ROI and feature mask are shown in Figure \ref{fig:implementation details}.

\section{Results and Discussion}
\subsection{Experimental Setup}

Three remote sensing image classification datasets across different modalities (RGB, multispectral and SAR) were used in this study: UCMerced \cite{yang2010bag},  EuroSAT \cite{helber2019eurosat} and Moving and Stationary Target Acquisition and Recognition (MSTAR) \cite{keydel1996mstar}. Data augmentation procedures were followed from \cite{peeples2021histogram} for UCMerced and EuroSAT, while MSTAR used only random crops \cite{chen2016target}. The experimental parameters for the models were the following: Adam optimization, initial learning rate of $.001$, batch size of $4$, and early stopping (patience of 5 and 10 for EuroSAT/UCMerced and MSTAR respectively). The number of epochs were 100 for MSTAR and 30 for UC Merced and EuroSAT. The pretrained models were applied, updating only the output layer to evaluate the performance of the feature extraction layers on each dataset.

For the XAI analysis, the methods and metrics were implemented using Captum \cite{kokhlikyan2020captum} and Quantus \cite{hedstrom2023quantus} respectively. Following \cite{bommer2023finding}, a subset of the test data is used to quantitatively evaluate the XAI methods for each dataset. The subset of samples is selected by ensuring that each class is represented equally. For example, in UCMerced, there are 21 classes. The number of samples per class is set to be ten, resulting in 210 examples to evaluate. EuroSAT and MSTAR were evaluated using a total of 100 and 40 examples respectively.

\subsection{Classification Performance}
\begin{table}[htb]
\begin{tabular}{|c|c|c|c|}
\hline
Model    & ConvNeXt                   & ViT              & FocalNets                 \\ \hline
UCMerced & 96.74 $\pm$  1.25          & 96.42 $\pm$ 0.58 & \textbf{96.90 $\pm$ 0.39} \\ \hline
EuroSAT  & \textbf{97.06  $\pm$ 0.00} & 92.56 $\pm$ 1.21 & 88.92 $\pm$ 1.03          \\ \hline
MSTAR    & \textbf{82.27 $\pm$ 3.53}  & 55.08  $\pm$  7.73 & 34.43 $\pm$ 6.16         \\ \hline
\end{tabular}
\caption{Average test classification accuracy ( $\pm$ 1 standard deviation) is shown for each model and dataset. The best average accuracy is bolded.}
\label{tab:Class}
\end{table}

\begin{figure*}[htb]
    \begin{subfigure}{.26\textwidth}{            \includegraphics[width=\textwidth]{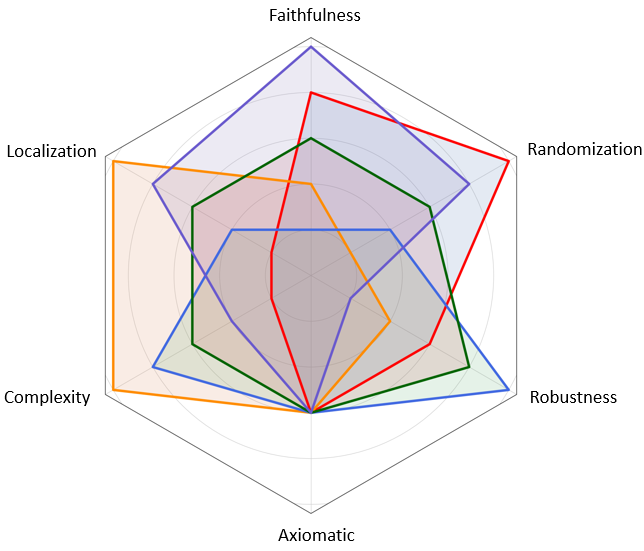}
            \caption{UCMerced}
            \label{fig:UCMerced}
        }
    \end{subfigure}
    \begin{subfigure}{.26\textwidth}{
            \includegraphics[width=\textwidth]{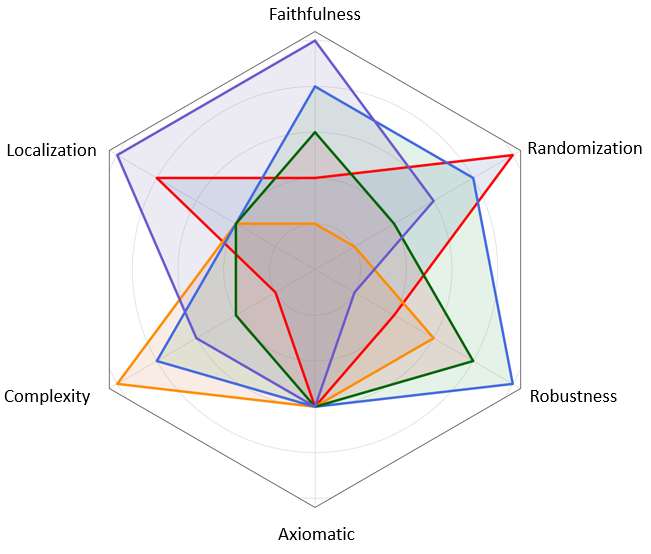}
            \caption{EuroSAT}
            \label{fig:EuroSAT}
        }
    \end{subfigure}
    \begin{subfigure}{.29\textwidth}
        \includegraphics[width=\textwidth]{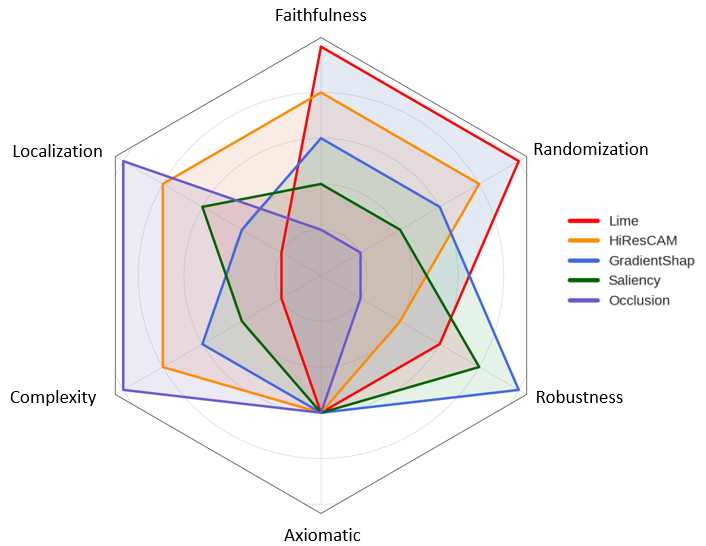}
        \caption{MSTAR}
        \label{fig:MSTAR}
    \end{subfigure}
    
    \begin{subfigure}{1.0\linewidth}{
            \includegraphics[width=\textwidth]{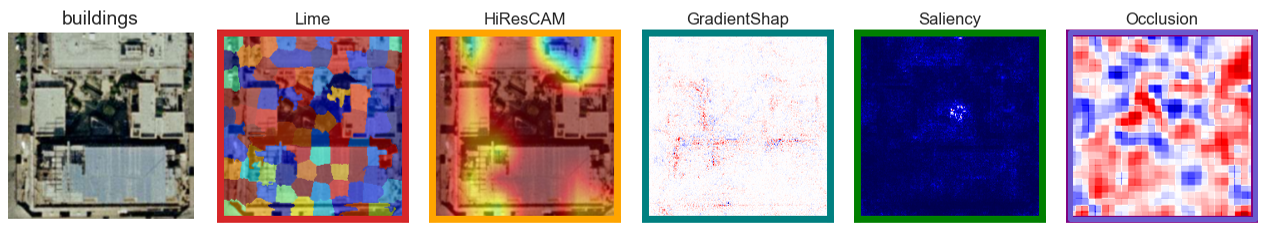}
            \caption{Example XAI Methods for UCMerced}
            \label{fig:XAI Outputs}
        }
    \end{subfigure}
   
    \caption{The average ranks across the three experimental runs for each XAI method on UCMerced using the ConvNeXt model is shown in the radar plots. The outer rank corresponds to the best XAI method (rank 1) according to the metric category and the center corresponding to the lowest ranked method (rank 5). Each color represents an XAI method as shown in Figure \ref{fig:XAI Outputs}.}

\end{figure*}

The classification results for each model on all three datasets is shown in Table \ref{tab:Class}. For the RGB and multispectral datasets (UCMerced and EuroSAT), each model performed well despite only tuning the output layer. However, the models did not perform as well on the MSTAR dataset. This results is not surprising as MSTAR has a limited number of specific target types and pretrained models tend to overfit on this dataset \cite{chen2016target}. ConvNeXt's features seem more generalizable as this model was fairly robust to the MSTAR dataset. 
\subsection{XAI Analysis}
\begin{table}[htb]
\resizebox{\textwidth}{!}{%
\begin{tabular}{|l|c|c|c|c|c|}
\hline
 & \multicolumn{1}{l|}{LIME} & \multicolumn{1}{l|}{HiResCAM} & \multicolumn{1}{l|}{GradientSHAP} & \multicolumn{1}{l|}{Saliency} & \multicolumn{1}{l|}{Occlusion} \\ \hline
Robustness    & 0.131 & 0.178 & \textbf{0.006} & 0.031 & 0.246 \\ \hline
Faithfulness  & 0.031 & 0.020  & 0.020  & 0.028 & \textbf{0.056} \\ \hline
Localization  & 0.392 & \textbf{0.437} & 0.398 & 0.412 & 0.413 \\ \hline
Complexity    & 0.401 & \textbf{0.652} & 0.566 & 0.445 & 0.443 \\ \hline
Randomization & \textbf{0.022} & {0.346} & 0.189 & 0.098 & 0.041 \\ \hline
Axiomatic     & 0     & 0     & 0     & 0     & 0     \\ \hline
\end{tabular}%
}
\caption{Normalized XAI scores for each XAI method on UCMerced dataset using the ConvNeXt model. The best average score per category is bolded. Standard deviations not included for brevity.}
\label{tab:UCMerced convnext}
\end{table}

The best classification model overall was ConvNeXt as this model achieved the highest average accuracy for EuroSAT and MSTAR while there was no significant difference between the UCMerced results as shown in Table \ref{tab:Class}. We further analyze ConvNeXt on UCMerced using the five primary attribution methods across the six different XAI metric categories in Table \ref{tab:UCMerced convnext}. The scores are normalized to account for whether metric should be minimized or maximized \cite{bommer2023finding}. One observation is that none of the methods satisfied the axiom constraint of completeness (similar results occurred using other axiomatic measures). New axiomatic measures could possibly be introduced to allow for some tolerance of error to quantify the extent to which an axiom is violated. 

The overall ranks across all three datasets using ConvNeXt is shown in Figures \ref{fig:EuroSAT} through \ref{fig:MSTAR}. Taking the average ranks of the metrics observed for the ConvneXt model, the Occlusion method performed best for the UCMerced dataset, GradientShap for the EuroSAT dataset, and HiResCAM performed the best for the MSTAR dataset. The Occlusion method iteratively substitutes image patches with a predetermined baseline and examines the model's predictions \cite{zeiler2013visualizing}. Specifically for the UCMerced dataset, the Occlusion method accurately represents the model's behavior (faithfulness), and the stability of model's output to perturbed inputs (randomization).
For EuroSAT, GradientShap's iterative substitution of a baseline value with randomly chosen subsets of attributions can successfully capture the links between image pixels and labels \cite{lundberg2017unified}. The attributions capture the relevant features and their impact on the model's predictions, providing a faithful explanation of the model's behavior.

The MSTAR dataset consist of targets centrally located in each image \cite{chen2016target}. HiResCAM generates attention maps that more accurately highlight the key areas for a given prediction \cite{draelos2021use}. As a result, the HiResCAM method will be the most well-rounded XAI approach (also supported by the metrics) to detect the contributions of the targets to the final prediction. The results vary across the different datasets for the other models. Certain measures of explainability may be more important depending on the application and end-users. An example qualitative result of the outputs from each method is shown in Figure \ref{fig:XAI Outputs}.


\section{Conclusion} We presented a benchmark study for quantitative analysis of XAI methods for remote sensing image classification. Future work includes evaluating our approach on other datasets (\textit{e.g.}, those that include geospatial information), additional analysis tasks (\textit{e.g.}, segmentation), finetuning every layer of each model, as well as developing new XAI methods and metrics. Also, XAI methods as well as metrics are dependent on the selection of the hyperparameters \cite{hedstrom2023quantus}. Future investigations can involve senstivity analysis of the selection of hyperparameters for both the XAI methods and metrics as this will impact additional analysis results.

\balance
\bibliographystyle{IEEEbib}
\bibliography{refs}

\begin{thebibliography}{10}

\bibitem{richards2022remote}
John~A Richards and John~A Richards,
\newblock {\em Remote sensing digital image analysis}, vol.~5,
\newblock Springer, 2022.

\bibitem{xu2022ai}
Yonghao Xu, Tao Bai, Weikang Yu, Shizhen Chang, Peter~M Atkinson, and Pedram
  Ghamisi,
\newblock ``Ai security for geoscience and remote sensing: Challenges and
  future trends,''
\newblock {\em arXiv preprint arXiv:2212.09360}, 2022.

\bibitem{hedstrom2023quantus}
Anna Hedstr{\"{o}}m, Leander Weber, Daniel Krakowczyk, Dilyara Bareeva, Franz
  Motzkus, Wojciech Samek, Sebastian Lapuschkin, and Marina Marina~M.{-}C.
  H{\"{o}}hne,
\newblock ``Quantus: An explainable ai toolkit for responsible evaluation of
  neural network explanations and beyond,''
\newblock {\em Journal of Machine Learning Research}, vol. 24, no. 34, pp.
  1--11, 2023.

\bibitem{bommer2023finding}
Philine Bommer, Marlene Kretschmer, Anna Hedstr{\"o}m, Dilyara Bareeva, and
  Marina M-C H{\"o}hne,
\newblock ``Finding the right xai method--a guide for the evaluation and
  ranking of explainable ai methods in climate science,''
\newblock {\em arXiv preprint arXiv:2303.00652}, 2023.

\bibitem{liu2022convnet}
Zhuang Liu, Hanzi Mao, Chao-Yuan Wu, Christoph Feichtenhofer, Trevor Darrell,
  and Saining Xie,
\newblock ``A convnet for the 2020s,''
\newblock in {\em Proceedings of the IEEE/CVF Conference on Computer Vision and
  Pattern Recognition}, 2022, pp. 11976--11986.

\bibitem{khan2022transformers}
Salman Khan, Muzammal Naseer, Munawar Hayat, Syed~Waqas Zamir, Fahad~Shahbaz
  Khan, and Mubarak Shah,
\newblock ``Transformers in vision: A survey,''
\newblock {\em ACM computing surveys (CSUR)}, vol. 54, no. 10s, pp. 1--41,
  2022.

\bibitem{yang2022focal}
Jianwei Yang, Chunyuan Li, Xiyang Dai, and Jianfeng Gao,
\newblock ``Focal modulation networks,''
\newblock {\em Advances in Neural Information Processing Systems (NeurIPS)},
  2022.

\bibitem{kakogeorgiou2021evaluating}
Ioannis Kakogeorgiou and Konstantinos Karantzalos,
\newblock ``Evaluating explainable artificial intelligence methods for
  multi-label deep learning classification tasks in remote sensing,''
\newblock {\em International Journal of Applied Earth Observation and
  Geoinformation}, vol. 103, pp. 102520, 2021.

\bibitem{papoutsis2023benchmarking}
Ioannis Papoutsis, Nikolaos~Ioannis Bountos, Angelos Zavras, Dimitrios Michail,
  and Christos Tryfonopoulos,
\newblock ``Benchmarking and scaling of deep learning models for land cover
  image classification,''
\newblock {\em ISPRS Journal of Photogrammetry and Remote Sensing}, vol. 195,
  pp. 250--268, 2023.

\bibitem{dosovitskiyimage}
Alexey Dosovitskiy, Lucas Beyer, Alexander Kolesnikov, Dirk Weissenborn,
  Xiaohua Zhai, Thomas Unterthiner, Mostafa Dehghani, Matthias Minderer, Georg
  Heigold, Sylvain Gelly, et~al.,
\newblock ``An image is worth 16x16 words: Transformers for image recognition
  at scale,''
\newblock in {\em International Conference on Learning Representations}, 2021.

\bibitem{kokhlikyan2020captum}
Narine Kokhlikyan, Vivek Miglani, Miguel Martin, Edward Wang, Bilal Alsallakh,
  Jonathan Reynolds, Alexander Melnikov, Natalia Kliushkina, Carlos Araya, Siqi
  Yan, and Orion Reblitz-Richardson,
\newblock ``Captum: A unified and generic model interpretability library for
  pytorch,'' 2020.

\bibitem{draelos2021use}
Rachel~Lea Draelos and Lawrence Carin,
\newblock ``Use hirescam instead of grad-cam for faithful explanations of
  convolutional neural networks,'' 2021.

\bibitem{lundberg2017unified}
Scott~M Lundberg and Su-In Lee,
\newblock ``A unified approach to interpreting model predictions,''
\newblock {\em Advances in neural information processing systems}, vol. 30,
  2017.

\bibitem{simonyan2014deep}
Karen Simonyan, Andrea Vedaldi, and Andrew Zisserman,
\newblock ``Deep inside convolutional networks: Visualising image
  classification models and saliency maps,'' 2014.

\bibitem{zhou2016learning}
Bolei Zhou, Aditya Khosla, Agata Lapedriza, Aude Oliva, and Antonio Torralba,
\newblock ``Learning deep features for discriminative localization,''
\newblock in {\em Proceedings of the IEEE conference on computer vision and
  pattern recognition}, 2016, pp. 2921--2929.

\bibitem{achanta2012slic}
Radhakrishna Achanta, Appu Shaji, Kevin Smith, Aurelien Lucchi, Pascal Fua, and
  Sabine S{\"u}sstrunk,
\newblock ``Slic superpixels compared to state-of-the-art superpixel methods,''
\newblock {\em IEEE transactions on pattern analysis and machine intelligence},
  vol. 34, no. 11, pp. 2274--2282, 2012.

\bibitem{otsu1979threshold}
Nobuyuki Otsu,
\newblock ``A threshold selection method from gray-level histograms,''
\newblock {\em IEEE transactions on systems, man, and cybernetics}, vol. 9, no.
  1, pp. 62--66, 1979.

\bibitem{yang2010bag}
Yi~Yang and Shawn Newsam,
\newblock ``Bag-of-visual-words and spatial extensions for land-use
  classification,''
\newblock in {\em Proceedings of the 18th SIGSPATIAL international conference
  on advances in geographic information systems}, 2010, pp. 270--279.

\bibitem{helber2019eurosat}
Patrick Helber, Benjamin Bischke, Andreas Dengel, and Damian Borth,
\newblock ``Eurosat: A novel dataset and deep learning benchmark for land use
  and land cover classification,''
\newblock {\em IEEE Journal of Selected Topics in Applied Earth Observations
  and Remote Sensing}, vol. 12, no. 7, pp. 2217--2226, 2019.

\bibitem{keydel1996mstar}
Eric~R Keydel, Shung~Wu Lee, and John~T Moore,
\newblock ``Mstar extended operating conditions: A tutorial,''
\newblock {\em Algorithms for Synthetic Aperture Radar Imagery III}, vol. 2757,
  pp. 228--242, 1996.

\bibitem{peeples2021histogram}
Joshua Peeples, Weihuang Xu, and Alina Zare,
\newblock ``Histogram layers for texture analysis,''
\newblock {\em IEEE Transactions on Artificial Intelligence}, vol. 3, no. 4,
  pp. 541--552, 2021.

\bibitem{chen2016target}
Sizhe Chen, Haipeng Wang, Feng Xu, and Ya-Qiu Jin,
\newblock ``Target classification using the deep convolutional networks for sar
  images,''
\newblock {\em IEEE transactions on geoscience and remote sensing}, vol. 54,
  no. 8, pp. 4806--4817, 2016.

\bibitem{zeiler2013visualizing}
Matthew~D Zeiler and Rob Fergus,
\newblock ``Visualizing and understanding convolutional networks,'' 2013.

\end{thebibliography}

\end{document}